\documentclass[10pt,twocolumn,letterpaper]{article}

\usepackage[pagenumbers]{cvpr} 
\usepackage{threeparttable}
\usepackage{tabularx}
\newcolumntype{Y}{>{\centering\arraybackslash}X}
\usepackage{graphicx}
\usepackage{booktabs}
\usepackage{color}
\usepackage{xcolor}
\usepackage{amssymb}
\usepackage{bm}
\usepackage{float}
\usepackage{wrapfig}
\usepackage{bbm}
\usepackage{mathrsfs}
\usepackage{subcaption}
\usepackage{multirow}
\usepackage{MnSymbol}
\usepackage{amssymb}
\usepackage{bm}
\usepackage{bbding}
\usepackage{pifont}

\makeatletter
\newcommand{\thickhline}{%
    \noalign {\ifnum 0=`}\fi \hrule height 1pt
    \futurelet \reserved@a \@xhline
}
\definecolor{mygray}{gray}{.9}
\definecolor{ggreen}{RGB}{18,190,108}
\usepackage{colortbl}

\usepackage{arydshln}
\usepackage{hhline}
\definecolor{cvprblue}{rgb}{0.21,0.49,0.74}
\usepackage[pagebackref,breaklinks,colorlinks,allcolors=cvprblue]{hyperref}

\title{HiFOR: Hierarchical Foresight with Action-Guided World Grounding for Vision-Language-Action Model}

\author{
Rongxing Ding\textsuperscript{1},
Hongyu Qu\textsuperscript{1,2},
Kehong Wu\textsuperscript{1},
Xudong Wang\textsuperscript{2,3},
Fan Zhang \textsuperscript{4},\\
Xinguang Xiang\textsuperscript{1}, 
Xiangbo Shu\textsuperscript{1}
\\[4pt]
\textsuperscript{1} Nanjing University of Science and Technology
\qquad
\textsuperscript{2} National University of Singapore \\
\textsuperscript{3} Shenyang Institute of Automation, Chinese Academy of Sciences\\
\textsuperscript{4} The Chinese University of Hong Kong \\
\vspace{-5pt}
}

\begin{document}
\maketitle
\begin{abstract}

Vision-Language-Action (VLA) models have demonstrated promising capabilities in robotic manipulation by integrating visual perception, language understanding, and action generation. Recent advances have further incorporated future prediction to enhance the understanding of environmental dynamics and task progression. However, existing methods often model short-term dynamics and long-term task objectives separately, while directly incorporating predicted futures into action policies without sufficiently considering their relevance to current action requirements. To address these limitations, we propose \textbf{HiFOR}, a hierarchical foresight framework with action-guided world retrieval. Specifically, \textbf{Hierarchical Foresight Prediction} jointly predicts short-term state evolution, semantic milestones, and final goal states through a Hierarchical Dependency Mask (HDM), establishing dependencies across temporal horizons. Building upon these predictions, \textbf{Structure-Aware Latent World Modeling} organizes future latent representations along the horizon, view, and spatial dimensions to facilitate fine-grained retrieval. Furthermore, \textbf{Action-Guided Factorized World Retrieval} adaptively retrieves relevant future information from the structured latent world based on evolving action representations during denoising, providing targeted guidance for action generation. Extensive experiments on two simulation benchmarks and six real-world robotic manipulation tasks demonstrate the effectiveness of our approach.

\end{abstract}    
\section{Introduction}
\label{sec:intro}
\begin{figure}[!t]
    \centering
    \includegraphics[width=0.99\linewidth]{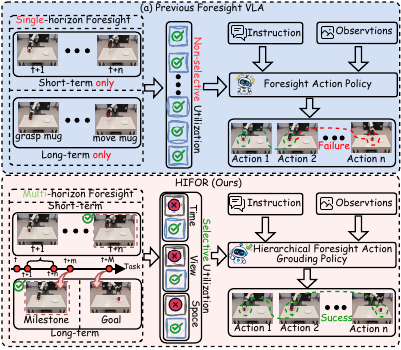}
    \caption{\textbf{Comparison of Foresight VLA Models.} (a) Existing methods typically adopt single-horizon foresight and directly feed predictions into VLA models without selective utilization. (b) In contrast, HiFOR combines multi-horizon foresight with action-guided retrieval of relevant future information.}
    \vspace{-20pt}
    \label{fig:motivation}
\end{figure}

Vision-Language-Action (VLA) models~\cite{kim2024openvla,intelligence2025pi_,yang2026eventvla,zheng2026xvla,shi2026memoryvla,luo2026being,zhu2026ati,liu2026plaw} have achieved remarkable progress in robotic manipulation by integrating visual perception, language understanding, and action generation within a unified end-to-end framework. Benefiting from advances in large-scale vision-language models and the growing availability of diverse robotic demonstrations, recent VLA models have demonstrated promising capabilities across a wide range of manipulation tasks. As robotic manipulation tasks become increasingly complex, anticipating future environmental changes and task evolution becomes essential for effective action generation. To this end, recent studies have explored incorporating foresight capabilities into VLA models to enhance their understanding of environmental dynamics and task progression~\cite{liu2026last,kim2026cosmos,xiao2026rehearsevla}. Existing approaches primarily predict future states or latent representations to model environmental dynamics~\cite{liu2026last,nguyen2026foca,zhang2026dreamvla,zheng2025flare,huang2025ladi}, or generate future images and videos to anticipate visual scene evolution~\cite{zhang2026foreact,kim2026cosmos,xiao2026rehearsevla,shen2026videovla,bu2024closed}, thereby providing additional foresight information for action generation.

Despite these advances, effectively translating future predictions into actionable guidance for complex robotic manipulation remains challenging. First, \textbf{modeling short-term world dynamics or long-term task objectives in isolation limits hierarchical coordination across temporal horizons.} As illustrated in Fig.~\ref{fig:motivation}(a), many existing foresight-based VLA methods primarily focus on a single prediction horizon. Short-term prediction methods~\cite{zhong2025flowvla,cen2025worldvla,yang2026mantis,li2026fomovla} emphasize near-future environmental changes and local motion dynamics, capturing fine-grained state evolution but often lacking explicit connections to subsequent task stages and ultimate goals. In contrast, long-term prediction methods~\cite{zhang2026foreact,chen2026goal,zhao2025cot,shou2026halo,long2026scaling} provide intermediate task guidance through semantic subtask planning or visual subgoal prediction, enabling the policy to anticipate key future task states. However, these methods often lack sufficient modeling of the continuous state transitions and motion dynamics required to achieve such objectives. Although both forms of foresight provide valuable information, modeling them independently makes it difficult to establish explicit dependencies among local state evolution, key semantic milestones, and final task objectives, thereby limiting the policy's coherent understanding of task progression across temporal horizons.

More importantly, \textbf{directly incorporating predicted future information into action generation often overlooks its varying relevance to current action requirements.} As illustrated in Fig.~\ref{fig:motivation}(a), many existing approaches~\cite{zhong2025flowvla,cen2025worldvla,yang2026mantis,zhang2026foreact,chen2026goal,zhao2025cot,wang2026wave} directly incorporate predicted future images or latent representations into action policies as additional conditioning signals, without sufficiently distinguishing the relevance of different predictions to current actions. However, future information across different prediction horizons, camera views, and spatial regions contributes differently to action generation, and its relevance dynamically evolves throughout task execution and the action generation process. Therefore, even with rich and complementary future predictions, selectively retrieving and utilizing information relevant to current action requirements remains essential for fully exploiting the decision-making potential of foresight. These observations motivate a unified framework that establishes hierarchical dependencies across temporal horizons and adaptively retrieves and leverages relevant future information according to current action requirements.

To address these challenges, we propose \textbf{HiFOR}, a hierarchical foresight framework with action-guided world retrieval for robotic manipulation. As illustrated in Fig.~\ref{fig:motivation}(b), HiFOR integrates multi-horizon foresight prediction with action-guided selective retrieval, enabling future information across different temporal horizons to provide targeted guidance according to evolving action requirements. Specifically, \ding{182} we first introduce \textbf{\textit{Hierarchical Foresight Prediction}}, which jointly predicts short-term state evolution, key semantic milestones, and final goal states. By incorporating a Hierarchical Dependency Mask (HDM), the prediction module establishes asymmetric dependencies across temporal horizons, connecting local motion dynamics with intermediate task progression and long-term objectives. \ding{183} Building upon these predictions, we propose \textbf{\textit{Structure-Aware Latent World Modeling}}, which organizes predicted future latent representations along the horizon, view, and spatial dimensions to construct a structured latent world while preserving fine-grained structural information for subsequent selective retrieval. \ding{184} Furthermore, we develop \textbf{\textit{Action-Guided Factorized World Retrieval}}, which leverages evolving action representations during denoising to perform factorized attention over the structured latent world. By adaptively evaluating the relevance of different prediction horizons, camera views, and spatial regions to current actions, the retrieved future information provides targeted guidance for action generation. Through these designs, HiFOR effectively integrates hierarchical future modeling with action-guided information retrieval, enabling foresight predictions to better support robotic manipulation.

Our main contributions are summarized as follows: \textbf{First,} we propose a \textbf{hierarchical foresight prediction mechanism} that jointly models short-term state evolution, key semantic milestones, and final goal states through hierarchical dependency modeling, establishing explicit connections between local world dynamics and long-term task objectives. \textbf{Second,} we develop an \textbf{action-guided structured world retrieval mechanism} that adaptively retrieves future information relevant to current action requirements from a structured latent world organized along the horizon, view, and spatial dimensions through denoising-aware factorized attention. \textbf{Third,} we conduct extensive experiments on LIBERO, RoboTwin 2.0, and six real-world robotic manipulation tasks, demonstrating the effectiveness of HiFOR and further validating the contributions of its key components through comprehensive ablation studies.
\section{Related Works}
\label{sec:formatting}

\subsection{Vision-Language-Action Model} 
\label{sec:related_vla}

Vision-Language-Action (VLA) models aim to integrate visual perception,
language understanding, and action generation to develop general-purpose
robotic policies capable of executing complex manipulation tasks.
Based on their action representations and generation mechanisms,
existing VLA models can be broadly categorized into two paradigms:
discrete and continuous action generation.

Discrete action generation methods~\cite{brohan2022rt,brohan2023rt,
kim2024openvla,ye2025latent,zhao2025cot,pertsch2025fast,
xu2026moeactok,liang2025discrete,yang2026chain}
convert continuous actions into discrete tokens or textual sequences,
enabling autoregressive action prediction.
In contrast, continuous action generation methods~\cite{kim2025fine,
black2024pi_0,intelligence2025pi_,li2026forcevla2,guo2026reflex}
typically employ specialized action prediction modules to directly
generate continuous control signals, mainly adopting regression-based,
diffusion-based, or flow-matching-based approaches.
Specifically, regression-based methods~\cite{kim2025fine,wu2024unleashing}
directly learn mappings from observations to actions,
while diffusion-based methods~\cite{liu2025rdt,wen2025diffusionvla,chuang2026lie,wen2025dexvla}
generate action sequences through iterative denoising.
Flow-matching-based methods~\cite{black2024pi_0,intelligence2025pi_,
routray2026vipra,zuo2026delta,tang2026fewer,shin2026dual} learn a continuous vector field to transform samples from a noise distribution into the target action distribution. Although these approaches have improved the action generation and generalization capabilities of VLA models, they primarily focus on action prediction itself, whereas accurate local action predictions do not necessarily guarantee successful completion of complex manipulation tasks.

\subsection{Foresight Predication in VLA } 
To overcome the limitations of local action prediction, recent studies have incorporated foresight prediction into VLA to anticipate future environmental changes and task progression, thereby guiding action generation. Based on how future information is represented, existing methods can be categorized into implicit and explicit foresight prediction. Implicit methods~\cite{liu2026last,nguyen2026foca,zhang2026dreamvla,zheng2025flare,huang2025ladi,yi2026placa} model environmental dynamics in state or latent feature spaces, leveraging predicted future representations to facilitate policy learning and action generation. In contrast, explicit methods~\cite{zhang2026foreact,kim2026cosmos,xiao2026rehearsevla,shen2026videovla,bu2024closed,cen2025rynnvla} generate future images or videos to anticipate visual scene evolution and potential action outcomes, supporting robotic planning and decision-making.

Despite these advances, existing methods still face limitations in modeling and utilizing future information. \textbf{First}, predicting short-term dynamics and long-term task objectives in isolation overlooks dependencies across temporal horizons, limiting coherent modeling of task progression. \textbf{Second}, existing methods often directly incorporate predicted future information into action policies as additional conditioning signals, without adequately considering its varying relevance to current action requirements.
\section{Method}
\begin{figure*}[!t]
    \centering
    \includegraphics[width=0.99\textwidth]{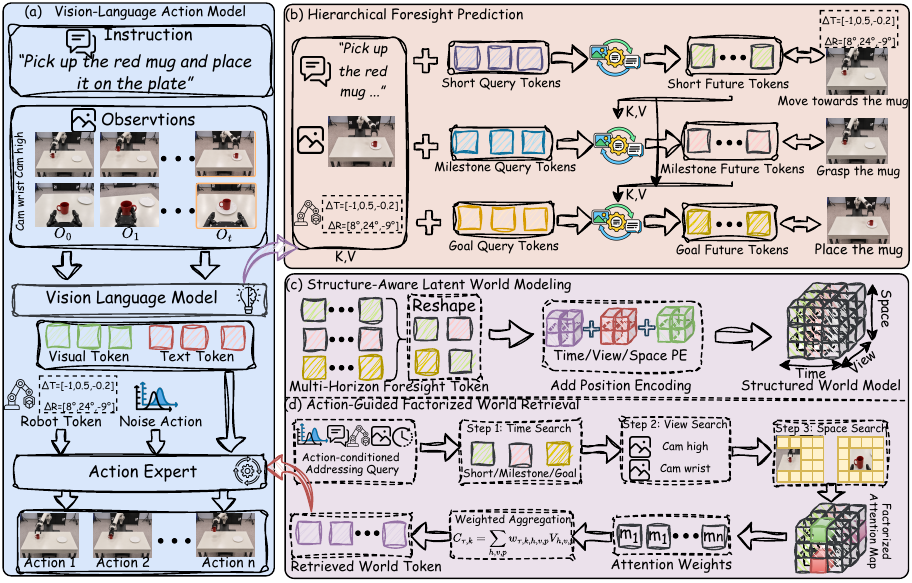}
    \caption{\textbf{The overview of HIFOR.} (a) We augment the VLA with hierarchical foresight prediction and action-guided world retrieval, enabling the Action Expert to leverage future information during action generation. (b) Hierarchical Foresight Prediction (\S\ref{sec::Foresight Prediction}) predicts short-term, milestone, and goal futures through hierarchical dependency modeling. (c) Structure-Aware Latent World Modeling (\S\ref{sec::Latent World}) organizes future latent representations along the horizon, view, and spatial dimensions. (d) Action-Guided Factorized World Retrieval (\S\ref{sec::World Retrieval}) selectively retrieves action-relevant future information through denoising-aware factorized attention to guide action generation.}
    \label{fig:overview}
    \vspace{-15pt}
\end{figure*}

\subsection{Problem Formulation}
\label{sec:problem}

We formulate robotic manipulation with VLA models as a language-conditioned sequential decision-making problem. 
At each timestep $t$, the policy receives multi-view visual observations 
$\mathcal{O}_t=\{\bm{I}_t^{v}\}_{v=1}^{V}$, a language instruction $\bm{L}$, and the robot state $\bm{s}_t$, and predicts a future action chunk:
\begin{equation}
\bm{A}_t
=
\{\bm{a}_{t+h}\}_{h=0}^{H-1}
=
\pi_{\theta}(\mathcal{O}_t,\bm{L},\bm{s}_t),
\end{equation}
where $H$ denotes the action horizon. Each action $\bm{a}_t =
\left[
\Delta x,\,
\Delta y,\,
\Delta z,\,
\Delta \theta_x,\,
\Delta \theta_y,\,
\Delta \theta_z,\,
g
\right]^{\top}$ consists of relative end-effector translation, rotation, and gripper control.

\subsection{Overview}

The overall framework of our method is illustrated in Fig.~\ref{fig:overview}. Given a language instruction, multi-view visual observations, and the robot state, we first employ a VLA backbone to encode the current multimodal context and generate action representations. We then introduce \textbf{\textit{Hierarchical Foresight Prediction}} (\S\ref{sec::Foresight Prediction}), which jointly predicts short-term state evolution, intermediate semantic milestones, and final goal states through a Hierarchical Dependency Mask (HDM), establishing dependencies across prediction horizons. Building upon these predictions, \textbf{\textit{Structure-Aware Latent World Modeling}} (\S\ref{sec::Latent World}) organizes the predicted future latent representations along the horizon, view, and spatial dimensions to construct a structured latent world model. Furthermore, we propose \textbf{\textit{Action-Guided Factorized World Retrieval}} (\S\ref{sec::World Retrieval}), which selectively retrieves action-relevant future information from the structured latent world to guide subsequent action generation. Together, these components enable our method to model hierarchical future evolution and selectively leverage relevant future information for action decision-making.

\subsection{Hierarchical Foresight Planning Dataset}
\begin{figure}[t]
    \centering
    \includegraphics[width=0.99\linewidth]{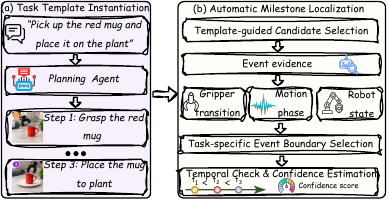}
    \vspace{-5pt}
    \caption{\textbf{Hierarchical Foresight Planning Data Construction.} (a) A planning agent generates semantic milestones from task instructions. (b) Automatic milestone localization identifies their completion timestamps in expert demonstrations using event evidence and temporal consistency checks.}
    \vspace{-15pt}
    \label{fig:dataset}
\end{figure}

We construct a hierarchical planning dataset from existing
successful expert demonstrations without collecting additional
robot trajectories. Instead of uniformly sampling future frames,
we identify semantically meaningful milestones to capture key
task progression across different temporal horizons,
providing structured future supervision for VLA training.

As illustrated in Fig.~\ref{fig:dataset}, we first employ a
Planning Agent to generate a task-specific semantic template
from the language instruction $\bm{L}$, decomposing it into
an ordered sequence of subgoals
$\mathcal{G}=\{G_1,G_2,\ldots,G_K\}$.
Each subgoal specifies a semantic stage, an event type,
and a completion verification criterion.
For example, ``Pick up the red mug and place it on the plate''
can be decomposed into \ding{182} \textit{grasp the mug},
\ding{183} \textit{move it to the target region}, and
\ding{184} \textit{place it on the plate},
with the number and types of milestones determined by the task.

Given the semantic template, we automatically localize the
completion timestamps of each subgoal in expert demonstrations.
Specifically, candidate event boundaries are identified using
gripper transitions, robot motion phases, and state information.
For \textbf{stable-state events}, such as grasping and placement,
we select the earliest frame where the target condition remains
persistently satisfied, while \textbf{dynamic events} are
localized based on motion or state-transition boundaries.
We further enforce temporal consistency
$t_1<t_2<\cdots<t_K$ and assign confidence scores to
the identified milestones, flagging ambiguous or low-confidence
cases for manual review.
The resulting annotations are represented as
$\mathcal{M}=\{(G_k,t_k)\}_{k=1}^{K}$.

To construct future visual supervision, we extract multi-view
semantic features from all trajectory frames using a frozen
DINOv2 ViT-L/14 encoder.
After removing prefix tokens, the patch features are
adaptively pooled into a $4\times4$ spatial grid:
\vspace{-5pt}
\begin{equation}
\bm{F}_t^v
=
\phi_{\mathrm{DINO}}(\bm{I}_t^v)
\in\mathbb{R}^{16\times1024}.
\vspace{-5pt}
\label{eq:dino_features}
\end{equation}
These features serve as visual supervision for subsequent
hierarchical foresight prediction.

\subsection{Hierarchical Foresight Prediction}
\label{sec::Foresight Prediction}
Complex robotic manipulation requires not only anticipating near-future environmental changes and motion dynamics but also understanding intermediate task progression and final goals. However, short-term world prediction and long-term semantic guidance in existing VLA methods often lack effective coordination, limiting their ability to capture task evolution across temporal scales. To address this, we propose \textbf{Hierarchical Foresight Prediction}. By establishing cross-horizon dependencies, our approach learns future representations that capture both local dynamics and long-term task objectives.

\noindent\textbf{Hierarchical Future Targets.}
Given the semantic milestone annotations
$\mathcal{M}=\{(G_k,t_k)\}_{k=1}^{K}$,
we define three complementary prediction horizons
for each timestep $t<t_K$.
Specifically, the short-term target is defined as
$t_t^s=\min(t+\Delta,t_K)$,
the milestone target as
$t_t^m=\min_{k:t_k>t}t_k$,
and the goal target as $t_t^g=t_K$,
corresponding to near-future state evolution,
the next semantic milestone, and the final task state,
respectively, where $\Delta$ denotes the short-term
prediction interval.

For each prediction horizon, we construct the corresponding
visual supervision using the precomputed multi-view DINOv2 features:
\begin{equation}
\bm{Z}_t^h[v,:,:]
=
\bm{F}_{t_t^h}^{v},
\quad
\bm{Z}_t^h\in\mathbb{R}^{V\times S\times D},
\quad h\in\{s,m,g\}.
\label{eq:future_targets}
\end{equation}
Here, $V$ denotes the number of camera views,
while $S=16$ and $D=1024$ represent the number of
spatial tokens per view and the feature dimension,
respectively.

Unlike long-term prediction, which focuses on intermediate semantic stages and final goal states, short-term prediction additionally requires modeling continuous robot motion during action execution. Since visual supervision at the target timestep alone may not sufficiently constrain fine-grained robot state changes, we introduce the future robot state sequence $\mathcal{S}_t^s =
(\bm{s}_{t+1},\bm{s}_{t+2},\ldots,\bm{s}_{t_t^s})$ as auxiliary supervision, where $\bm{s}_t$ denotes the robot state at timestep $t$. Thus, short-term prediction combines visual and state
supervision to capture local motion dynamics, while long-term prediction models task progression and final goal states through future visual supervision.

\noindent\textbf{Hierarchical Future Prediction.}
To predict the hierarchical future targets, we encode the current multi-view observations $\mathcal{O}_t$ and language instruction $\bm{L}$ into visual tokens $\bm{V}_t$ and language tokens $\bm{E}_L$ using the VLM. We introduce three groups of learnable future queries, $\bm{Q}^s$, $\bm{Q}^m$, and $\bm{Q}^g$, corresponding to short-term, milestone, and goal prediction. To establish cross-horizon dependencies, we design a
\textbf{Hierarchical Dependency Mask (HDM)} that progressively
expands attention visibility from short-term to goal queries.
Under HDM, future queries at different temporal horizons
interact hierarchically through a shared prediction module,
with the prediction process formulated as:
\begin{equation}
\bm{U}_t^h =
\mathcal{H}_{\theta}
\left(
\bm{Q}^h;
[\bm{V}_t;\bm{E}_L;\bm{Q}^{\preceq h}]
\right),
\quad h\in\{s,m,g\}.
\label{eq:hierarchical_queries}
\end{equation}
Here, $\mathcal{H}_{\theta}$ denotes the shared prediction
module, and $\bm{Q}^{\preceq h}$ represents the concatenation
of query groups from the short-term level up to the current
horizon $h$, following the hierarchical order $s\prec m\prec g$.
All query groups are jointly updated within the shared network
under HDM, producing the corresponding hierarchical prediction
features $\bm{U}_t^h$.

We project the predicted features into the DINOv2 feature space using visual prediction heads:
\begin{equation}
\hat{\bm{Z}}_t^h =
\mathcal{P}_{\mathrm{vis}}^h(\bm{U}_t^h)
\in \mathbb{R}^{V\times S\times D},
\quad h\in\{s,m,g\}.
\label{eq:future_visual_prediction}
\end{equation}
For short-term prediction, an additional state prediction head estimates the future robot state sequence $\hat{\mathcal{S}}_t^s =
\mathcal{P}_{\mathrm{state}}(\bm{U}_t^s)$.
The overall foresight prediction loss is defined as
\begin{equation}
\begin{aligned}
\mathcal{L}_{\mathrm{HFP}}
=
\lambda_{m}
\mathcal{D}_{\mathrm{v}}
\left(\hat{\bm{Z}}_t^h,\bm{Z}_t^h\right)
+
\lambda_{\mathrm{n}}
\mathcal{D}_{\mathrm{s}}
\left(\hat{\mathcal{S}}_t^s,\mathcal{S}_t^s\right),
\end{aligned}
\label{eq:foresight_loss}
\end{equation}
where $\mathcal{D}_{\mathrm{vis}}$ and $\mathcal{D}_{\mathrm{state}}$ denote the visual and robot state prediction losses, respectively, and $\lambda_m$ and $\lambda_{n}$ are the corresponding weights.

\subsection{Structure-Aware Latent World Modeling}
\label{sec::Latent World}
The Hierarchical Foresight Prediction module generates future visual representations at three temporal horizons: short-term, milestone, and goal. These predictions contain complementary information across temporal horizons, camera views, and spatial regions. However, organizing these predictions into a unified token sequence without explicitly preserving their structural attributes hinders the effective utilization of future information. To address this, we propose \textbf{Structure-Aware Latent World Modeling}, which organizes multi-horizon predictions into a Structured Latent World, explicitly preserving their horizon, view, and spatial structures to facilitate subsequent fine-grained information retrieval.

Specifically, given the predicted future features $\{\hat{\bm{Z}}_t^h\}_{h\in\{s,m,g\}}$, we construct structured world tokens through feature projection and structural positional embeddings:
\vspace{-5pt}
\begin{equation}
\bm{w}_t^{h,v,p}
=
\mathcal{P}_{\mathrm{w}}
\left(\hat{\bm{Z}}_t^h[v,p,:]\right)
+
\bm{e}_h^{\mathrm{hor}}
+
\bm{e}_v^{\mathrm{view}}
+
\bm{e}_p^{\mathrm{spa}},
\vspace{-5pt}
\label{eq:world_tokens}
\end{equation}
where $\mathcal{P}_{\mathrm{w}}$ denotes the feature projection, and $\bm{e}_h^{\mathrm{hor}}$, $\bm{e}_v^{\mathrm{view}}$, and $\bm{e}_p^{\mathrm{spa}}$ represent horizon, view, and spatial positional embeddings, respectively. Here, $p\in\{1,\ldots,S\}$ indexes the spatial tokens within each camera view.

We then organize these tokens into a unified structured latent world representation:
\vspace{-5pt}
\begin{equation}
\bm{W}_t[h,v,p,:]
=
\bm{w}_t^{h,v,p},
\quad
\bm{W}_t \in \mathbb{R}^{3\times V\times S\times d_w}.
\vspace{-5pt}
\label{eq:structured_world}
\end{equation}
where $d_w$ denotes the latent world feature dimension. This representation explicitly preserves the horizon, view, and spatial dimensions, enabling fine-grained access to predicted future information.

To support world representations at different granularities, we further derive horizon-level and view-level features through spatial aggregation:
\vspace{-5pt}
\begin{equation}
\bm{r}_t^h
=
\frac{1}{VS}
\sum_{v=1}^{V}
\sum_{p=1}^{S}
\bm{w}_t^{h,v,p},
\bm{r}_t^{h,v}
=
\frac{1}{S}
\sum_{p=1}^{S}
\bm{w}_t^{h,v,p}.
\vspace{-5pt}
\label{eq:horizon_representation}
\end{equation}
Here, $\bm{r}_t^h$ summarizes the predicted world state at horizon $h$, while $\bm{r}_t^{h,v}$ captures view-specific information within that horizon. The original world tokens $\bm{w}_t^{h,v,p}$ retain fine-grained spatial details. Together, these representations provide horizon, view, and spatial information, facilitating subsequent action-guided retrieval of task-relevant future features.

\subsection{Action-Guided Factorized World Retrieval}
\label{sec::World Retrieval}

After constructing the structured latent world, we further introduce \textbf{\textit{Action-Guided Factorized World Retrieval (AWR)}}. Since the relevance of future information to current action requirements may vary across denoising steps, we construct dynamic retrieval queries from the action expert's hidden representations at each denoising step. We then perform factorized attention along the horizon, view, and spatial dimensions to selectively retrieve future information relevant to the ongoing action generation process.

Specifically, given the structured latent world
$\bm{W}_t\in\mathbb{R}^{3\times V\times S\times d_w}$,
we construct an action-conditioned query
$\bm{q}_{t,\tau}^{a}=\mathcal{Q}(\bm{H}_{t,\tau}^{a})$
from the action expert's hidden representations
$\bm{H}_{t,\tau}^{a}$ at denoising step $\tau$.
Three independent linear projections then generate queries
corresponding to temporal horizons, camera views, and spatial
locations, denoted as $\bm{q}_{t,\tau}^{H}$,
$\bm{q}_{t,\tau}^{V}$, and $\bm{q}_{t,\tau}^{S}$,
respectively.
Meanwhile, we employ a shared linear projection to map
the hierarchical features of the structured latent world into
corresponding key representations $\bar{\bm{k}}_t^h$,
$\bar{\bm{k}}_t^{h,v}$, and $\bm{k}_t^{h,v,p}$,
while projecting the spatial-level world tokens into value
representations $\bm{u}_t^{h,v,p}$ for subsequent factorized
attention-based retrieval.

Based on these queries and keys, we formulate a unified
factorized attention operation across the horizon, view,
and spatial dimensions:
\begin{equation}
a_{t,\tau}^{\ell}(i\mid c)
=
\operatorname{Softmax}_{i}
\left(
\frac{
(\bm{q}_{t,\tau}^{\ell})^\top
\bm{k}_t^{\ell}(i\mid c)
}{\sqrt{d}}
\right),
\quad \ell\in\{H,V,S\}.
\label{eq:factorized_attention}
\end{equation}
Here, $\ell\in\{H,V,S\}$ denotes the horizon, view,
and spatial attention levels, with the corresponding
index-condition pairs $(i,c)$ given by
$(h,\varnothing)$, $(v,h)$, and $(p,(h,v))$,
respectively.
The associated key representations are
$\bar{\bm{k}}_t^h$, $\bar{\bm{k}}_t^{h,v}$,
and $\bm{k}_t^{h,v,p}$, where $d$ denotes
the feature dimension per attention head.
The resulting attention weights are denoted as
$\alpha_{t,\tau}^{h}$, $\beta_{t,\tau}^{v\mid h}$,
and $\gamma_{t,\tau}^{p\mid h,v}$, respectively.

We then combine the three attention weights to aggregate
the latent world values, yielding the action-relevant
future representation at denoising step $\tau$:
\begin{equation}
\bm{f}_{t,\tau}
=
W_O
\left(
\sum_{h\in\{s,m,g\}}
\sum_{v=1}^{V}
\sum_{p=1}^{S}
\alpha_{t,\tau}^{h}
\beta_{t,\tau}^{v\mid h}
\gamma_{t,\tau}^{p\mid h,v}
\bm{u}_t^{h,v,p}
\right).
\label{eq:factorized_world_retrieval}
\end{equation}
Here, $W_O$ denotes the output projection.
As the action-conditioned queries evolve throughout denoising,
the model dynamically adjusts its attention across different
prediction horizons, camera views, and spatial regions.
Finally, the retrieved future features are fused with
the action representations and short-term robot state features,
and injected into the action expert through residual connections
to guide iterative action generation.

\noindent\textbf{Overall Loss Function.} HiFOR jointly optimizes action generation and hierarchical
foresight prediction through the flow-matching loss $\mathcal{L}_{\mathrm{ACT}}$ and the foresight prediction loss $\mathcal{L}_{\mathrm{HFP}}$, respectively. The overall training objective is defined as:
\begin{equation}
\mathcal{L} = \mathcal{L}_{\mathrm{ACT}} + \mathcal{L}_{\mathrm{HFP}}.
\end{equation}

\section{Experiments}

\subsection{Evaluation on Simulation Benchmarks}

\noindent\textbf{Simulation Benchmarks.} To evaluate the robustness of our method under diverse simulated environments, we perform experiments on both LIBERO~\cite{liu2023libero} and RoboTwin 2.0~\cite{chen2025robotwin}. For LIBERO~\cite{liu2023libero}, we report the average success rate over four task suites, including Spatial, Object, Goal, and Long, where each suite contains 10 tasks. Each task suite is evaluated with 500 rollouts to obtain reliable performance statistics. For RoboTwin 2.0~\cite{chen2025robotwin}, we follow the Hard evaluation protocol and assess our method on 20 randomly selected tasks with 100 rollouts per task. The Hard setting introduces challenging domain randomizations, including variations in clutter, illumination, textures, and object heights.

\noindent\textbf{Implementation Details.} All experiments are conducted on 8 $\times$ H200 GPUs with a global batch size of 512 (64 per GPU). Both LIBERO~\cite{liu2023libero} and RoboTwin 2.0~\cite{chen2025robotwin} are trained end-to-end for 30000 optimization steps using AdamW. The learning rate is linearly warmed up to $5\times10^{-5}$ over the first 5000 steps and then kept constant for the remaining training. We apply gradient clipping with a maximum norm of 1.0 and maintain an exponential moving average of the model parameters with a decay rate of 0.999.

\begin{table*}[!t]
    \centering
    \caption{
    \textbf{Performance comparison on RoboTwin 2.0~\cite{chen2025robotwin}}. We report per-task success rates (SR) over 100 rollouts under the \textit{Hard} setting. $\dagger$ denotes reproduced results. Results for all 20 tasks are provided in the Appendix.
    }
    \label{tab:robotwin_hard}
    \vspace{-5pt}
    \setlength{\tabcolsep}{5.0pt}
    \renewcommand{\arraystretch}{1.05}
    \resizebox{\textwidth}{!}{
    \begin{tabular}{l||ccccccccccc}
        \thickhline
        \rowcolor{mygray}
        & \multicolumn{11}{c}{\textbf{RoboTwin 2.0 (Hard)}} \\
        \hhline{~||-----------}
        \rowcolor{mygray}
        & Adjust & Blocks & Click & Dump Bin & Move & Open & Place Burger & Place Can & Place Cans & Stack Blocks & \\
        \rowcolor{mygray}
        \multirow{-3}{*}{\textbf{Method}} & Bottle & Ranking RGB & Bell & Bigbin & Playingcard & Laptop & Fries & Basket & Plasticbox & Two & \multirow{-2}{*}{\textbf{Avg.}} \\
        \hline
        RDT~\cite{liu2025rdt} & 75\% & 0\% & 9\% & 32\% & 11\% & 32\% & 27\% & 6\% & 5\% & 2\% & 14.60\% \\
        ACT~\cite{zhao2023learning} & 23\% & 0\% & 3\% & 1\% & 0\% & 0\% & 0\% & 0\% & 0\% & 0\% & 2.90\% \\
        DP3~\cite{ze20243d} & 3\% & 0\% & 0\% & 53\% & 3\% & 7\% & 18\% & 2\% & 3\% & 0\% & 4.85\% \\
        EventVLA~\cite{yang2026eventvla} & 24\% & 6\% & 48\% & 64\% & 17\% & 23\% & 7\% & 7\% & 14\% & 13\% & 19.00\% \\
        Spatial Forcing~\cite{li2026spatial} & 71\% & 7\% & 28\% & 70\% & 15\% & 15\% & 71\% & 10\% & 11\% & 19\% & 29.40\% \\
        X-VLA~\cite{zheng2026xvla} & 33\% & 4\% & 63\% & 41\% & 5\% & 16\% & 35\% & 9\% & 60\% & 2\% & 23.25\% \\
        $\pi_{0}$~\cite{black2024pi_0} & 56\% & 5\% & 3\% & 24\% & 22\% & 46\% & 4\% & 5\% & 2\% & 1\% & 14.60\% \\
        $\pi_{0.5}^{\dagger}$~\cite{intelligence2025pi_} & 77\% & 29\% & 35\% & 67\% & 40\% & 59\% & 82\% & 4\% & 40\% & 32\% & 40.35\% \\
        \hline
        \textbf{HiFOR (Ours)} & \textbf{83\%} & \textbf{46\%} & \textbf{48\%} & \textbf{75\%} & \textbf{54\%} & \textbf{65\%} & \textbf{88\%} & \textbf{14\%} & \textbf{63\%} & \textbf{45\%} & \textbf{49.40\%} \\
        \hline
    \end{tabular}
    }
    \vspace{-10pt}
\end{table*}

\begin{table}[!t]
    \centering
    \caption{\textbf{Success rates (\%) on the LIBERO benchmark}.
    Best results are shown in \textbf{bold}, and second-best results are \underline{underlined}.}
    \vspace{-5pt}
    \label{tab:libero_results}
    \resizebox{\linewidth}{!}{
    \begin{tabular}{lc||ccccc}
        \thickhline
        \rowcolor{mygray}
        \textbf{Method}
        &  \textbf{Publication} & \textbf{Spatial} & \textbf{Object} & \textbf{Goal}& \textbf{Long} & \textbf{Avg.} \\
        \hline
        Diffusion Policy~\cite{chi2025diffusion} & IJRR'25 & 78.5 & 87.5 & 73.5 & 64.8 & 76.1 \\
        GR00T-N1~\cite{bjorck2025gr00t} & arXiv'25 & 94.4 & 97.6 & 93.0 & 90.6 & 93.9 \\
        OpenVLA~\cite{kim2024openvla} & arXiv'24 & 84.7 & 88.4 & 79.2 & 53.7 & 76.5 \\
        OpenVLA-OFT~\cite{kim2025fine} & arXiv'25 & 97.6 & 98.4 & 97.9 & 94.5 & 97.1 \\
        MemoryVLA~\cite{shi2026memoryvla} & ICLR'26 & 98.4 & 98.4 & 96.4 & 93.4 & 96.7 \\
        UniVLA~\cite{bu2025univla} & arXiv'25 & 95.4 & 98.8 & 95.8 & 85.2 & 94.2 \\
        $\pi_0$-Fast~\cite{pertsch2025fast} & AAAI'26 & 96.4 & 96.8 & 88.6 & 60.2 & 85.5\\
        $\pi_0$~\cite{black2024pi_0} & arXiv'24 & 96.8 & 98.8 & 95.8 &85.2 & 94.2 \\
        $\pi_{0.5}$~\cite{intelligence2025pi_} & arXiv'25 & \underline{98.8} & 98.2 & \underline{98.0} & 92.4 & 96.9 \\
        VLA-Adapter~\cite{wang2026vla} & AAAI'26 & 97.8 & 99.2 & 97.2 & \underline{95.0} & \underline{97.3} \\
        Fast-ThinkAct~\cite{huang2026fast} & CVPR'26 & 92.0 & 97.2 & 90.2 & 79.4 & 89.7 \\
        LEAP-VLA~\cite{yuleap} & ECCV'26 & 98.4 & \textbf{99.4} & 97.6 & 93.0 & 97.1 \\
         \hline
        \textbf{HiFOR (Ours)} & - & \textbf{99.4} & \underline{99.2} & \textbf{98.6} & \textbf{95.8} & \textbf{98.3} \\
        \hline
    \end{tabular}
    }
    \vspace{-15pt}
\end{table}

\noindent\textbf{Performance on RoboTwin 2.0 Benchmark.} As shown in Table~\ref{tab:robotwin_hard}, HiFOR achieves the best overall performance on RoboTwin 2.0 under the Hard setting, reaching an average success rate of \textbf{49.40\%} across 20 tasks, outperforming $\pi_{0.5}$ by \textbf{9.05\%}. Specifically, HiFOR improves the success rates on Blocks Ranking RGB and Place Cans Plasticbox from 29\% and 40\% to \textbf{46\%} and \textbf{63\%}, respectively. Moreover, on Move Playingcard and Stack Blocks Two, HiFOR achieves success rates of \textbf{54\%} and \textbf{45\%}, surpassing the baseline by \textbf{14\%} and \textbf{13\%}, respectively. These results demonstrate the effectiveness of HiFOR in improving performance across diverse and challenging robotic manipulation tasks.

\noindent\textbf{Performance on LIBERO Benchmark.} As shown in Table~\ref{tab:libero_results}, HiFOR achieves the best overall performance on LIBERO, with an average success rate of \textbf{98.3\%}. Specifically, it achieves the highest success rates of \textbf{99.4\%}, \textbf{98.6\%} and \textbf{95.8\%} on LIBERO-Spatial, LIBERO-Goal and LIBERO-Long, respectively, while remaining highly competitive on other task suites. Compared with the $\pi_{0.5}$ baseline, HiFOR improves the average success rate from 96.9\% to \textbf{98.3\%}, with the largest gain of \textbf{3.4\%} on long-horizon tasks. These results demonstrate that HiFOR effectively enhances long-horizon task execution while maintaining strong performance across diverse manipulation tasks.

\subsection{Evaluation on Real-World}

\noindent\textbf{Training and Evaluation Setting.}
To evaluate the effectiveness of HiFOR in real-world robotic manipulation, we conduct experiments on \textbf{\textit{Generalization Tasks}} and \textbf{\textit{Long-horizon Tasks}} using an AgileX Piper 6-DoF single-arm manipulator. The robot is equipped with a parallel-jaw gripper and two Intel RealSense D435i RGB cameras, providing wrist-mounted and third-person visual observations. For Generalization Tasks, we use 100 expert demonstrations per task for training, while Long-horizon Tasks use 200 demonstrations per task. All models are trained on 8 $\times$ H200 GPUs and deployed on a single NVIDIA RTX 4090 GPU for real-world evaluation. Both task categories are evaluated with 50 rollouts per task, with object positions randomly initialized in each rollout to assess robustness and generalization under varying initial conditions.

\begin{figure*}[!t]
    \centering
    \includegraphics[width=0.99\linewidth]{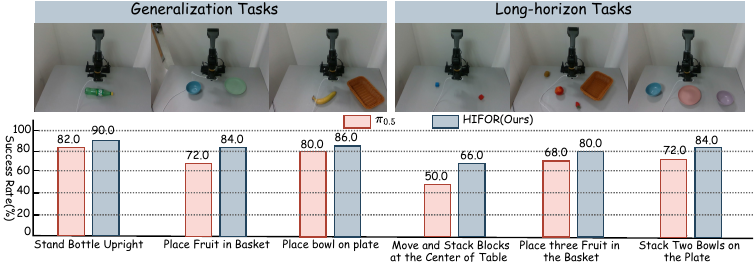}
    \vspace{-8pt}
    \caption{\textbf{Real-world evaluation results}.We evaluate HIFOR against leading models on the Generalization and Long-horizon task suites.}
    \vspace{-15pt}
    \label{fig:real_task}
\end{figure*}

\noindent\textbf{Performance on Real-World Setup.} To evaluate the effectiveness of HiFOR in real-world robotic manipulation, we conduct experiments on six tasks involving object grasping, placement, and stacking. As shown in Fig.~\ref{fig:real_task}, HiFOR consistently outperforms $\pi_{0.5}$ across all tasks, achieving an average success rate of \textbf{81.7}\%, an improvement of \textbf{11.0}\% over $\pi_{0.5}$. Notably, HiFOR achieves a \textbf{16.0}\% improvement on the challenging block stacking task, demonstrating its effectiveness in complex real-world manipulation.

\begin{table}[!t]
    \centering
    \caption{
   \textbf{Ablation analysis of the core components} in terms of Success Rate (\%) on LIBERO~\cite{liu2023libero} and RoboTwin 2.0~\cite{chen2025robotwin}.  
    }
    \vspace{-5pt}
    \label{tab:ablation1}
    \setlength{\tabcolsep}{5.0pt}
    \renewcommand{\arraystretch}{1.05}
    \resizebox{\linewidth}{!}{
    \begin{tabular}{cccc||cccc||c}
    \thickhline
    \rowcolor{mygray}
    \multicolumn{4}{c||}{\textbf{Method}} &\multicolumn{4}{c||}{\textbf{LIBERO}} &\textbf{RoboTwin2.0} \\
    \rowcolor{mygray}
    Short & Milestone & Goal & AWR & Spatial & Object & Goal & Long & Avg. \\
    \hline
     & & & & 98.8 & 98.2 & 98.0 & 92.4 & 40.35 \\
    $\checkmark$ & & & & 99.0 & 98.4 & 98.0 & 93.4 &44.15  \\
    $\checkmark$ & $\checkmark$ & & & 99.2 & 99.0 & 98.0 & 94.2 &45.60  \\
    $\checkmark$ &$\checkmark$ & $\checkmark$ & & 99.0 & 98.8 & 98.2 & 95.0 & 46.95  \\
    \arrayrulecolor{gray}\hdashline\arrayrulecolor{black}
    $\checkmark$ &$\checkmark$ & $\checkmark$ & $\checkmark$ & 99.4 & 99.2 & 98.6& 95.8& 49.40 \\
    \hline
    \end{tabular}
    }
    \vspace{-15pt}
\end{table}

\subsection{Ablation Study}
\begin{figure}[!t]
    \centering
    \includegraphics[width=0.95\linewidth]{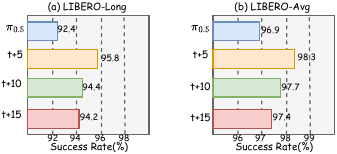}
    \vspace{-5pt}
    \caption{\textbf{Analysis of the Effects of Different Short-Horizon Prediction Lengths} on LIBERO.~\cite{liu2023libero}.}
    \vspace{-10pt}
    \label{fig:ablation2}
\end{figure}

\noindent\textbf{Key Component Analysis.} We evaluate the key components of HiFOR through ablation studies in Table~\ref{tab:ablation1}. Introducing \textbf{short-term foresight prediction} improves the average success rate on RoboTwin 2.0 from 40.35\% to \textbf{44.15\%}, while incorporating \textbf{milestone} and \textbf{goal} predictions further increases it to \textbf{45.60\%} and \textbf{46.95\%}, respectively. Adding \textbf{action-guided world retrieval (AWR)} further raises the success rate to \textbf{49.40\%}, an additional gain of \textbf{2.45}\%. Consistent improvements are also observed on LIBERO, particularly LIBERO-Long, where the success rate increases from 92.4\% to \textbf{95.8\%}. These results highlight the complementary benefits of hierarchical foresight prediction and action-guided retrieval in effectively leveraging future information for robotic manipulation.

\noindent\textbf{Different Short-horizon Prediction Length.}
We investigate the effect of different short-horizon prediction lengths on LIBERO, as illustrated in Fig.~\ref{fig:ablation2}. Specifically, we evaluate prediction horizons of 5, 10, and 15 steps. The 5-step horizon achieves success rates of \textbf{95.8\%} and \textbf{98.3\%} on LIBERO-Long and LIBERO-Avg, respectively. Notably, the policy updates its predictions every 5 execution steps on LIBERO, naturally aligning the 5-step prediction horizon with the action execution interval. This alignment enables short-term foresight to provide effective guidance within the current execution cycle, whereas longer prediction horizons extend beyond the next prediction update. 

\begin{figure}[!t]
    \centering
    \includegraphics[width=0.99\linewidth]{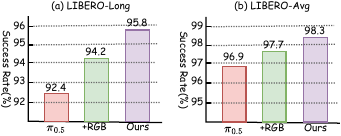}
    \vspace{-5pt}
    \caption{\textbf{Analysis of different future prediction representations} on LIBERO~\cite{liu2023libero}.}
    \vspace{-15pt}
    \label{fig:ablation3}
\end{figure}

\noindent\textbf{Different Future Prediction Representations.} We investigate the impact of different future prediction targets by comparing direct pixel-level prediction of future RGB images with semantic feature-based future prediction. As shown in Fig.~\ref{fig:ablation3}, RGB-based prediction improves success rates from 92.4\% to 94.2\% on LIBERO-Long and from 96.9\% to 97.7\% on LIBERO-Avg. Our method further achieves \textbf{95.8\%} and \textbf{98.3\%}, outperforming RGB-based prediction by \textbf{1.6} and \textbf{0.6} percentage points, respectively. These results demonstrate the benefits of future-aware supervision and suggest that semantic feature-based prediction provides more effective guidance for action generation, particularly in long-horizon manipulation tasks.

\subsection{Qualitative Analysis} 
\begin{figure}[!t]
    \centering
    \includegraphics[width=1\linewidth]{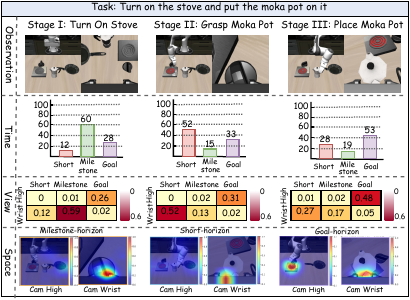}
    \vspace{-15pt}
    \caption{\textbf{The Visualization of Action-Guided Factorized World Retrieval} Across Task Stages.}
    \vspace{-15pt}
    \label{fig:spatio}
\end{figure}

\noindent\textbf{Hierarchical Spatiotemporal Attention Visualization.} To investigate how our method utilizes structured future information, we visualize the hierarchical spatiotemporal attention at representative timesteps from three task stages. As shown in Fig.~\ref{fig:spatio}, during stove activation, the model primarily attends to the milestone horizon (60\%), focusing on the wrist view and stove control knob. During grasping, attention shifts toward the short-term horizon (52\%), emphasizing local manipulation details from the wrist view. During final placement, the goal horizon receives the highest attention (53\%), with the third-person view highlighting the target placement region. These results illustrate that our method adaptively retrieves task-relevant future information across temporal horizons, camera views, and spatial regions.
\section{Conclusion}
In this paper, we propose HiFOR, a hierarchical foresight-guided framework for vision-language-action robotic manipulation. HiFOR employs Hierarchical Foresight Prediction to jointly predict short-term dynamics, intermediate semantic milestones, and final goal states, capturing dependencies across different prediction horizons. Structure-Aware Latent World Modeling further organizes these predictions along the horizon, view, and spatial dimensions to construct a structured latent world. Action-Guided Factorized World Retrieval then adaptively retrieves action-relevant future information to guide subsequent action generation. Extensive experiments on LIBERO and RoboTwin 2.0, together with real-world evaluations, demonstrate the effectiveness of HiFOR in improving robotic manipulation performance.
{
    \small
    \bibliographystyle{ieeenat_fullname}
    \bibliography{main}
}


\end{document}